\documentclass[runningheads]{llncs}
\usepackage[T1]{fontenc}
\usepackage{graphicx}
\usepackage{comment}
\usepackage{pgfplots}
\pgfplotsset{compat=1.16}
\usepackage{xurl}
\usepackage{xcolor}
\usepackage{booktabs}
\usepackage{cite}
\usepackage{hyperref}
\usepackage{amsmath}
\usepackage{amssymb}
\usepackage{siunitx}

\DeclareMathOperator*{\argmin}{arg\,min}

\begin{document}
\title{$\operatorname{cIDIR}$: Conditioned Implicit Neural Representation for Regularized Deformable Image Registration}

\titlerunning{$\operatorname{cIDIR}$}
%



\author{
Sidaty El hadramy \thanks{Corresponding author. E-mail: \email{sidaty.elhadramy@unibas.ch}} \and 
Oumeymah Cherkaoui \and
Philippe C. Cattin}

\institute{Department of Biomedical Engineering, University of Basel, Switzerland}

\authorrunning{El Hadramy et al.}

\maketitle              
\begin{abstract}

Regularization is essential in deformable image registration (DIR) to ensure that the estimated Deformation Vector Field (DVF) remains smooth, physically plausible, and anatomically consistent. However, fine-tuning regularization parameters in learning-based DIR frameworks is computationally expensive, often requiring multiple training iterations. To address this, we propose $\operatorname{cIDIR}$, a novel DIR framework based on Implicit Neural Representations (INRs) that conditions the registration process on regularization hyperparameters. Unlike conventional methods that require retraining for each regularization hyperparameter setting, $\operatorname{cIDIR}$ is trained over a prior distribution of these hyperparameters, then optimized over the regularization hyperparameters by using the segmentations masks as an observation. Additionally, $\operatorname{cIDIR}$ models a continuous and differentiable DVF, enabling seamless integration of advanced regularization techniques via automatic differentiation. Evaluated on the DIR-LAB \cite{Castillo2009-ko, Castillo2010-ji} dataset, $\operatorname{cIDIR}$ achieves high accuracy and robustness across the dataset.

\end{abstract} 
\section{Introduction}\label{sec1}

Deformable image registration (DIR) is essential in medical imaging for aligning images across different views, modalities, time points, or patients. It enables image fusion and analysis, supporting applications in diagnosis, treatment planning, and intervention guidance \cite{Maintz1998-wr, El_hadramy2023-wl}. To enhance the effectiveness of DIR, enforcing a diffeomorphic transformation is critical \cite{Vercauteren2007-nx}, as it ensures physical plausibility while minimizing artifacts and distortions that could affect medical image interpretation. Theoretically, a diffeomorphic transformation $\phi$ is both smooth and invertible \cite{Glaunes2008-fg}. To enforce these properties, various regularization strategies have been proposed. Rohlfing \textit{et al.} \cite{Rohlfing2003-yr} introduced a Jacobian regularization, where a negative determinant of the Jacobian matrix indicates a loss of invertibility. Burger \textit{et al.} \cite{Burger2013-jv} proposed a hyperelastic regularization term to control variations in length, surface area, and volume. Alvarez \textit{et al.} \cite{Alvarez2024-vs} introduced a regularization term enforcing the conservation of linear momentum. Additionally, Rueckert \textit{et al.} \cite{Rueckert1999-pk} developed a Bending Energy regularization to further ensure the smoothness of the deformation vector field (DVF).

Recent advances in deep neural networks have led to the introduction of numerous DIR approaches \cite{Fu2020-fz}. These methods, supervised \cite{Lafarge2018-ga} or unsupervised \cite{Balakrishnan2019-md}, learn to predict DVF on a grid for unseen image pairs. However, these grid-based methods provide a discontinuous representation of the DVF, making it challenging to incorporate advanced regularization techniques that require accurate gradient computations. In particular, regularization methods that rely on second-order gradient calculations face challenges in this context. Therefore, the need to incorporate these regularization techniques has driven the development of methods aimed at learning continuous and differentiable representations of the DVF. One class of method that has gained attention relies on implicit neural representations (INRs) \cite{Molaei2023-ub}. INRs use Multi-Layer Perceptrons (MLP) to encode information as a continuous generator function, mapping input coordinates to corresponding values within the defined space.

Building on INRs, Wolterink \textit{et al.} \cite{Wolterink2022-nv} introduced $\operatorname{IDIR}$, an implicit deformable image registration model that seamlessly integrates regularization techniques, facilitated by automatic differentiation techniques. While the method, validated on the DIR-LAB 4DCT dataset \cite{Castillo2009-ko,Castillo2010-ji}, demonstrated high accuracy when using the Bending Energy regularization \cite{Rueckert1999-pk}, it requires hyperparameter tuning to balance the weights between data and regularization losses. Standard hyperparameter optimization methods, such as random search, grid search, and sequential search \cite{Bergstra2012-gz}, are commonly used for this aim. More advanced techniques, including gradient-based tuning and Bayesian optimization, use probabilistic models to efficiently identify optimal values \cite{Mockus1974-xq,Snoek2012-hj}. These methods are computationally intensive, as they require retraining the model multiple times to assess each hyperparameter choice. To avoid this, Hoopes \textit{et al.} \cite{Hoopes2021-yq} introduced HyperMorph, which uses a hypernetwork to condition the main registration network (VoxelMorph \cite{Balakrishnan2019-md}) on the loss hyperparameters. Thus, the hyperparameter tuning can be done at inference-time. However, training hypernetworks is often challenging due to the high difference between the network's input and output dimensions, which leads to slow convergence and high memory consumption \cite{Ortiz2023-sj}. 

In this work, we build upon $\operatorname{IDIR}$ by introducing $\operatorname{cIDIR}$, a simple, yet effective approach that conditions an INR of the DVF on the hyperparameters of the loss functions. Similar to $\operatorname{IDIR}$, $\operatorname{cIDIR}$ offers a continuous and differentiable representation of the DVF, making it easier to integrate advanced regularization techniques. The key difference, however, lies in how hyperparameter tuning is handled. While $\operatorname{IDIR}$ requires multiple training sessions to optimize the hyperparameters, $\operatorname{cIDIR}$ only needs to be trained once. Hyperparameter tuning is performed after training, allowing for real-time adjustments. This makes $\operatorname{cIDIR}$ more efficient and practical for applications where quick, on-the-fly registration is necessary. The paper is organized as follows: Section~\ref{Methods} introduces $\operatorname{cIDIR}$ and outlines its novelty. In Section~\ref{Experiments}, we evaluate $\operatorname{cIDIR}$ on the DIR-LAB 4DCT dataset \cite{Castillo2009-ko,Castillo2010-ji}, demonstrating its accuracy and computational efficiency. Finally, Section~\ref{Conclusion} concludes the paper and discusses future research directions.

\section{Methods} \label{Methods}

This section presents the proposed approach, starting with the formulation of the deformable image registration problem and key notations. We then describe the \boldmath{$\operatorname{cIDIR}$} architecture and its components, followed by a discussion of the regularization techniques used in this work and its hyperparameters optimization.

\subsection{Deformable Image Registration}

Let $\mathcal{M}$ and $\mathcal{F}$ be the moving and fixed images, respectively, defined over the domains $\Omega_0 \subset \mathbb{R}^d$ and $\Omega \subset \mathbb{R}^d$. The goal of deformable image registration is to estimate a DVF \boldmath{$\phi$} that aligns $\mathcal{M}$ with $\mathcal{F}$ by minimizing a given criterion. As in Equation~\ref{eq:optimization_problem}, this is formulated as an optimization problem. Where $\phi$ is sought to map features from the moving image to their corresponding homologous structures in the fixed image while preserving anatomical consistency.

\begin{equation}
    \hat{\phi} = \argmin_{\phi} {(1 - \alpha) \mathcal{L}_{sim} (\mathcal{M} \circ \phi, \mathcal{F}) + \alpha \mathcal{L}_{reg}(\phi)}
    \label{eq:optimization_problem}
\end{equation}

\noindent In Equation~\ref{eq:optimization_problem}, $\phi$ is a DVF, $\mathcal{L}_{sim}$ denotes the similarity measure, $\mathcal{L}_{reg}$  is the regularization penalty, and $\alpha$ is the weighting factor for regularization. In this work, \textbf{we treat \boldmath{$\alpha$} as the hyperparameter to be tuned}, assuming a predefined similarity measure and a fixed regularization technique.

\subsection{$\operatorname{cIDIR}$}

\begin{figure}[!h]
    \centering
    \includegraphics[width=0.7\linewidth]{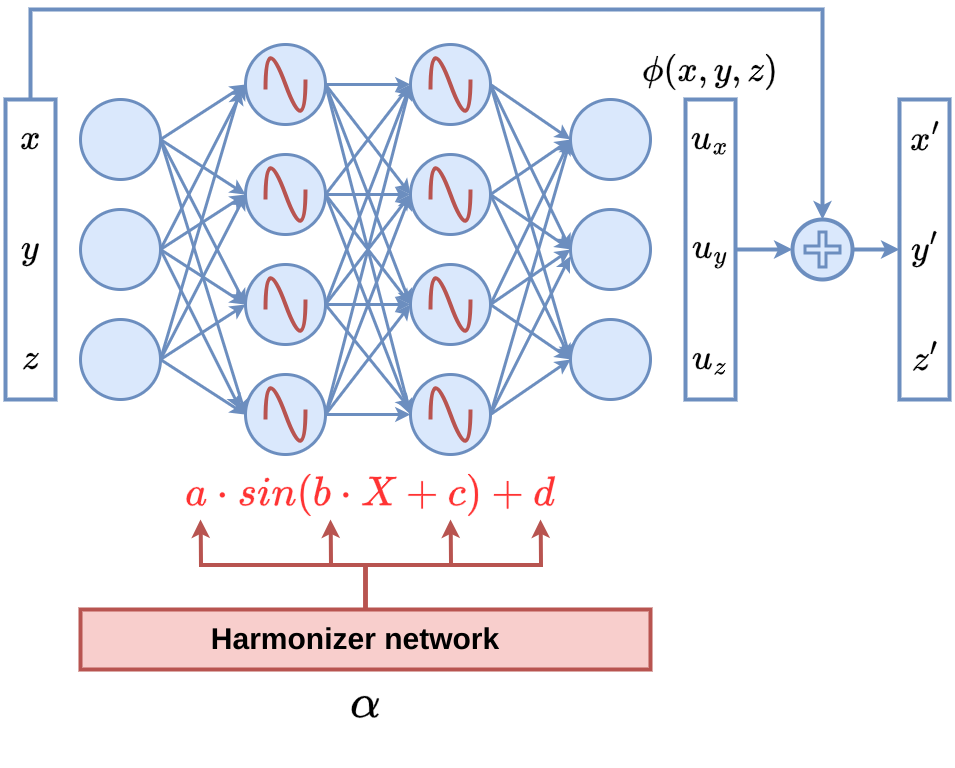}
    \caption{Overview of $\operatorname{cIDIR}$. The main network (in blue) learns an implicit representation of the Deformation Vector Field (DVF) $\phi$, mapping coordinates $(x,y,z)$ from the moving image to $(x',y', z')$ in the fixed image. The network is conditioned on the regularization weighting factor $\alpha$ through the harmonization network, which predicts the parameters $a$, $b$, $c$, and $d$ of the activation function used in the main network.}
    \label{fig:cidir-overview}
\end{figure}

To address the optimization problem outlined in the previous section, we introduce a learning-based framework called $\operatorname{cIDIR}$. Figure~\ref{fig:cidir-overview} shows an overview of the framework. $\operatorname{cIDIR}$ consists of two key components: the main network and the harmonizer network. The \textbf{main network} learns an implicit representation of the DVF $\phi$, while the \textbf{harmonizer network} conditions it on the regularization weighting factor \boldmath{$\alpha$}. Both networks are trained in an end-to-end manner, with \boldmath{$\alpha$} uniformly sampled from $[0, 1]$ during training. 

The \textbf{main network} in $\operatorname{cIDIR}$ is a MLP with an input and output dimension of 3, corresponding to the spatial coordinates of the moving and fixed images. It consists of three hidden layers, each with 256 neurons, followed by an activation function. Since the network is designed as an INR, the choice of activation function is critical. Standard activations like ReLU are unsuitable, as they tend to bias the network toward low-frequency signals \cite{anonymous2024where}. To address this, prior works have explored the use of periodic activation functions \cite{Wolterink2022-nv}, which enable the network to capture high-frequency variations effectively. In this work, we employ a parameterized activation function $\sigma$, inspired by \cite{kazerouni2024incode}, which is formulated as:
\begin{equation}
    \sigma(x) = \boldmath{a \cdot \operatorname{sin}(b \cdot x + c) + d}
    \label{eq:activation_function}
\end{equation}

\noindent In Equation~\ref{eq:activation_function}, the parameters of this activation function play distinct roles in shaping the output response of the network. \boldmath{$a$} represents the amplitude scaling, determining the vertical scaling of the sinusoidal function. \boldmath{$b$} represents the frequency scaling, influencing how rapidly the function oscillates. \boldmath{$c$} represents the phase shift, it adjusts the horizontal displacement of the function along the $x$-axis. \boldmath{$d$} is the vertical shift of the wave, acting as a baseline adjustment. By learning these parameters, the activation function becomes highly adaptive, enabling the network to model complex deformations.

The \textbf{harmonizer network} is an MLP designed to condition the main network on the regularization weighting factor $\alpha$. It takes $\alpha$ as input and outputs four values corresponding to the parameters $a, b, c, d$ of the main network's activation function. The network consists of three hidden layers with sizes $128$, $64$, and $32$, respectively. Each hidden layer is followed by \textbf{layer normalization} to stabilize training and improve generalization, as well as a SiLU (Sigmoid Linear Unit) activation function, which enhances smooth gradient propagation and avoids vanishing gradients. The choice of the hidden layer dimensions was determined experimentally to balance model accuracy and computational efficiency. By dynamically predicting the activation function parameters, the harmonizer network enables adaptive control over the deformation field, allowing for a conditioning of the prediction over the weighting factor $\alpha$.


\subsection{Regularizations}
This section briefly presents three common regularization techniques in DIR to ensure a smooth, physically plausible DVF. $\operatorname{cIDIR} $'s implicit representation enables seamless integration of these techniques, efficiently computing gradients of different orders for advanced regularization.

\noindent \textbf{Jacobian Regularization}, formulated in Equation \ref{eq:jacobian_regularization}, ensures a diffeomorphic DVF, the Jacobian determinant must remain non-negative at each point $x$, as negative values indicate non-invertibility. Therefore, as proposed by \cite{Rohlfing2003-yr}, deviations from $1$ are minimized to maintain a stable and realistic transformation.

\begin{equation}
    \mathcal{L}_{jac}(\phi) = \int_{\Omega} |\det \nabla \phi - 1 |\,dx
    \label{eq:jacobian_regularization}
\end{equation}

\noindent \textbf{Hyperelastic Regularization} \cite{Burger2013-jv} fromulated in Equation~\ref{eq:hyperelastic_regularization}. It regulates the length, surface area, and volume of the DVF. The Jacobian matrix governs length, while the cofactor matrix and determinant control area and volume. 

\begin{equation}
\mathcal{L}^{\text{hyper}}[\phi] = \int_{\Omega} \left[ \frac{1}{2} \alpha_l \left| \nabla u \right|^2 + \alpha_a \phi_c (\text{cof} \, \nabla \phi) + \alpha_v \psi (\det \nabla \phi) \right] dx
\label{eq:hyperelastic_regularization}
\end{equation}

\noindent \textbf{Bending Energy Penalty} \cite{Rueckert1999-pk} enforces the smoothness of the DVF by penalizing large second derivatives, ensuring that the DVF remains smooth across the entire domain. This is formulated as:

\begin{equation}
  \begin{aligned}
    \mathcal{L}^{\text{bending}}[\phi] = & \frac{1}{8} \int_{-1}^{1} \int_{-1}^{1} \int_{-1}^{1} \left[ \left( \frac{\partial^2 \phi}{\partial x^2} \right)^2 + \left( \frac{\partial^2 \phi}{\partial y^2} \right)^2 + \left( \frac{\partial^2 \Phi}{\partial z^2} \right)^2 \right. \\
    &\left. + 2 \left( \frac{\partial^2 \phi}{\partial xy} \right)^2 + 2 \left( \frac{\partial^2 \phi}{\partial xz} \right)^2 + 2 \left( \frac{\partial^2 \phi}{\partial yz} \right)^2 \right] dx\, dy\, dz.
    \end{aligned}
\end{equation}

\subsection{Optimization of the regularization weights}\label{sec:opt}

Upon $\operatorname{cIDIR}$'s training, a grid search over values of $\alpha$ in the range [0, 1] is performed. For each $\alpha$, a displacement field is generated and applied to the moving image to produce a \textbf{moved image}. Both the \textbf{moved and fixed images} are then segmented using the method from \cite{Hofmanninger2020-zq}, and the resulting segmentations are used to compute a Dice score (DS). The $\alpha$ value that yields the highest DS is selected as the optimal. In this setup, \textbf{the segmentations serve as observations to guide the optimization of} $\alpha$.

\section{Results} \label{Experiments}
\subsection{Dataset and implementation details}

We evaluate our method on the DIR-LAB 4DCT dataset \cite{Castillo2009-ko,Castillo2010-ji}, a widely used benchmark in DIR \cite{Wolterink2022-nv, Fechter2020-wv, Hering2021-um}. This dataset consists of 4D CT scans from $10$ patients, where the registration task involves aligning inspiration-phase images with expiration-phase images. The challenge arises due to the combined effects of cardiac and respiratory motion, which cause significant deformations.  For each patient, the dataset provides landmark coordinates in the fixed image along with their corresponding positions in the moving image. All networks are implemented in PyTorch, and experiments are conducted on an NVIDIA A100 GPU with 40GB of memory.  $\operatorname{cIDIR}$ is trained using the Adam optimizer with a learning rate of $10^{-4}$. The Bayesian Optimization ($\operatorname{BO}$) is implemented with Scikit-Optimize and the Grid Search ($\operatorname{GS}$) with Scikit-Learn.
 
\subsection{Experiments}
For each of the $10$ patients in the DIR-LAB 4DCT dataset, $\operatorname{cIDIR}$ was trained for 50K epochs, with the regularization weight $\alpha$ varying between $0$ and $1$ using Bending Energy regularization. In each epoch, a batch of 10K points was sampled from the lung region of the moving image, identified using a segmentation mask computed with the method from Hofmanninger \textit{et al.} \cite{Hofmanninger2020-zq}. These point coordinates, along with a sampled $\alpha$ value, were input into $\operatorname{cIDIR}$ to predict their corresponding positions in the fixed image. $\operatorname{cIDIR}$ is trained by minimizing the voxel intensity differences between the predicted and actual points in the fixed image using the Normalized Cross-Correlation (NCC) as $\mathcal{L}_{sim}$ loss in Equation~\ref{eq:optimization_problem}. After training, the optimal value of $\alpha$ is obtained using the approach described in \ref{sec:opt}. Table~\ref{tab:results} presents the results, comparing $\operatorname{cIDIR}$ using Bending Energy regularization with state-of-the-art learning-based methods, including $\operatorname{IDIR}$ \cite{Wolterink2022-nv}, and $\operatorname{CNN}$ \cite{Fechter2020-wv}. Additionally, we report the initial \textbf{Displacement} error. For $\operatorname{IDIR}$, we performed experiments using a Bending Energy regularization with $\alpha = 10$, as suggested in the original paper \cite{Wolterink2022-nv}. When reproducing $\operatorname{IDIR}$'s results, we observed improved performance for patients $01$, $02$, $03$, $04$, and $06$ with $\alpha = 10$ compared to the results reported in their paper. However, for patients $05$, $07$, $08$, $09$, and $10$, we were unable to match the accuracy reported in their work. Table~\ref{tab:results} demonstrates that $\operatorname{cIDIR}$, by selecting the optimal $\alpha$ in real-time during inference, outperforms $\operatorname{IDIR}$ and $\operatorname{CNN}$ \cite{Fechter2020-wv} in terms of average Target Registration Error (TRE) over the $10$ patients. This is because $\operatorname{cIDIR}$ allows for real-time grid search to find the optimal $\alpha$ for each patient after training. In contrast, the fixed $\alpha = 10$ used in $\operatorname{IDIR}$ might not be optimal for all patients, and optimizing it is time-consuming, requiring multiple training.


\begin{table}[!h]
    \centering
    \caption{TRE in \SI{}{mm} of $\operatorname{cIDIR}$ compared to state-of-the-art learning-based methods: $\operatorname{IDIR}$ \cite{Wolterink2022-nv}, $\operatorname{CNN}$ \cite{Fechter2020-wv}, and $\operatorname{Displacement}$ (TRE before registration). Both $\operatorname{IDIR}$ and $\operatorname{cIDIR}$ use Bending Energy regularization, with $\alpha = 10$ for $\operatorname{IDIR}$ as proposed in their paper. For $\operatorname{cIDIR}$, the value of $\alpha$ is optimized per patient.}
    \label{tab:results}
    \begin{tabular}{lccccc}
        \toprule
        Scan & $\operatorname{cIDIR}$ \textbf{(ours)} & $\operatorname{IDIR}$ \cite{Wolterink2022-nv}  & $\operatorname{CNN}$ \cite{Fechter2020-wv} & $\operatorname{Displacement}$\\
        \midrule
        4DCT 01 & 0.66 (1.25) & 0.52 (1.11) &  1.21 (0.88) & 4.01 (2.91) \\
        4DCT 02 & 0.76 (1.33) & 0.55 (1.15) &  1.13 (0.65) & 4.65 (4.09) \\
        4DCT 03 & 0.68 (1.23) & 0.76 (1.32) &  1.32 (0.82) &  6.73 (4.21) \\
        4DCT 04 & 1.18 (1.3) & 0.82 (1.47) &  1.84 (1.76) &  9.42 (4.81) \\
        4DCT 05 & 1.17 (1.86) & 1.29 (1.78) &  1.80 (1.60) &  7.10 (5.14) \\
        4DCT 06 & 0.82 (1.84) & 0.86 (1.40) &  2.30 (3.78) &  11.10 (6.98) \\
        4DCT 07 & 1.35 (1.65) & 1.76 (2.29) &  1.91 (1.65) &  11.59 (7.87) \\
        4DCT 08 & 1.44 (3.05) & 2.54 (4.30) &  3.47 (5.00)  & 15.16 (9.11) \\
        4DCT 09 & 3.72 (2.59) & 3.54 (2.65) &  1.47 (0.85) & 7.82 (3.99) \\
        4DCT 10 & 1.61 (2.14) & 1.50 (1.94) &  1.79 (2.24) &  7.63 (6.54) \\
        \midrule
        \textbf{Average} & \textbf{1.33} & 1.47 & 1.83 &  8.52 \\
        \bottomrule
    \end{tabular}
\end{table}

To further evaluate $\operatorname{cIDIR} $'s adaptability across different regularization techniques, we conducted experiments using both hyperelasticity and Jacobian regularizations. For $\operatorname{IDIR}$, we trained the model with a fixed value of $\alpha = 0.5$ across all $10$ patients. $\operatorname{cIDIR}$ was trained on a uniform distribution of $\alpha$ over $[0, 1]$, and at inference, we selected $\alpha = 0.5$. As shown in Table~\ref{tab:results_different_regularizations}, both methods achieved comparable results for the Jacobian regularization, while $\operatorname{cIDIR}$ outperformed $\operatorname{IDIR}$ with hyperelasticity. This difference can be attributed to the fact that the optimal $\alpha$ value is not necessarily the same for both methods, even when using the same regularization technique. Since $\operatorname{cIDIR}$ incorporates a harmonizer network that conditions the registration process on $\alpha$, its architecture is more complex than $\operatorname{IDIR}$. This added complexity not only allows $\operatorname{cIDIR}$ to generalize over a range of $\alpha$ values but also affects how the regularization influences the learned deformation field, potentially leading to different optimal $\alpha$ for both methods.

\begin{table}[!ht]
\centering
\caption{Comparison of TRE in \SI{}{\mm} for $\operatorname{cIDIR}$ and $\operatorname{IDIR}$ on the DIR-LAB \cite{Castillo2009-ko,Castillo2010-ji} dataset using Hyperelastic \cite{Burger2013-jv} and Jacobian \cite{Rohlfing2003-yr} regularizations. $\operatorname{IDIR}$ is trained with a fixed $\alpha = 0.5$, while $\operatorname{cIDIR}$ is trained over $\alpha \in [0,1]$ and evaluated with $\alpha = 0.5$.}\label{tab:results_different_regularizations}
\begin{tabular*}{0.8\textwidth}{@{\extracolsep\fill}c c c c c }
\toprule
Scan & \multicolumn{2}{c}{\textbf{Hyperelastic} $(\alpha = 0.5)$ } & \multicolumn{2}{c}{\textbf{Jacobian} $(\alpha = 0.5)$} \\
               & $\operatorname{cIDIR}$  & $\operatorname{IDIR}$ & $\operatorname{cIDIR}$  & $\operatorname{IDIR}$   \\
\toprule
4DCT 01  & 0.73 (1.16) & 7.10 (5.48) & 1.46 (1.77)  &  1.61 (2.18)  \\
4DCT 02  & 0.63 (1.14) & 3.34 (3.09) & 2.78 (2.31)  & 2.55 (3.06) \\
4DCT 03  & 1.32 (1.80) & 3.99 (2.81) & 4.11 (2.78)  &  3.17 (3.59) \\
4DCT 04  & 1.11 (1.44) & 7.83 (4.90) & 5.65 (4.65)  &  3.95 (5.72) \\
4DCT 05  & 1.42 (1.79) & 6.35 (5.36) & 2.86 (3.13)  & 2.29 (2.99) \\
4DCT 06  & 1.08 (1.35) & 8.70 (6.64)  & 3.68 (3.32)  &  5.95 (6.99) \\
4DCT 07  & 2.48 (2.48) & 6.97 (7.37) & 6.40 (5.32)  &  6.20 (5.18) \\
4DCT 08  & 6.00 (5.71) & 7.89 (7.98) & 11.4 (10.9)  &  11.01 (11.51) \\
4DCT 09  & 3.59 (3.10) & 7.23 (6.02) & 3.97 (2.41) &  3.15 (2.49) \\
4DCT 10  & 2.17 (2.09) & 5.17 (4.52) & 6.91 (4.4)  &  6.26 (5.34) \\
\midrule
\textbf{Average} & 2.05 & 6.45  & 4.92   & 4.61 \\
\bottomrule
\end{tabular*}
\end{table}

\subsection{Computation Time}

$\operatorname{cIDIR}$ offers significant benefits in terms of computation time. Table~\ref{tab:time} presents training and fine-tuning time per regularization technique for both $\operatorname{cIDIR}$ and $\operatorname{IDIR}$ over a patient.  While a single $\operatorname{IDIR}$ training is faster than $\operatorname{cIDIR}$ due to its shorter training duration, $\operatorname{cIDIR}$ is training longer (50K epochs) with $\alpha$ varying over a uniform distribution \([0,1]\). However, after $\operatorname{cIDIR}$'s training, $\alpha$ is optimized in near real-time, requiring only a single prediction per evaluation, which is performed in about \SI{1}{ms}. This allows a \textbf{Grid Search} ($\operatorname{GS}$) over multiple values of $\alpha$ in less than two seconds (see Table~\ref{tab:time}). In contrast, $\operatorname{IDIR}$ requires a full training run for each $\alpha$ evaluation. To optimize $\operatorname{IDIR}$ over $\alpha$, we employed \textbf{Bayesian Optimization} ($\operatorname{BO}$) with a Gaussian Process as a surrogate model to efficiently search for the best $\alpha$ within \([0,1]\). Instead of exhaustively testing all values, the Gaussian Process models the performance landscape, guiding the search towards promising regions and reducing the number of required training runs. However, as shown in Table~\ref{tab:time}, at least $20$ training iterations of $\operatorname{IDIR}$ were needed for convergence to an optimal $\alpha$, making its fine-tuning process longer than the total training and fine-tuning time of $\operatorname{cIDIR}$.

\begin{table}[!ht]
\centering
\caption{Training and fine-tuning time per regularization technique for $\operatorname{cIDIR}$ and $\operatorname{IDIR}$ on 4DCT 04. We use a \textbf{Bayesian Optimization} ($\operatorname{BO}$) to fine-tune $\alpha$ for $\operatorname{IDIR}$, which requires multiple full training runs. In contrast, $\operatorname{cIDIR}$ selects the optimal $\alpha$ using a simple \textbf{Grid Search} ($\operatorname{GS}$) (See Section \ref{sec:opt})}

\label{tab:time}
\begin{tabular*}{\textwidth}{@{\extracolsep{\fill}}c c c c c c c c c}
\toprule
Scan & \multicolumn{2}{c}{Hyperelastic \cite{Burger2013-jv}} & \multicolumn{2}{c}{Jacobian \cite{Rohlfing2003-yr}} & \multicolumn{2}{c}{Bending \cite{Rueckert1999-pk}} \\
     & Training & Fine-tuning & Training & Fine-tuning & Training & Fine-tuning \\
\midrule
$\operatorname{IDIR}$ \cite{Wolterink2022-nv}  $+ \operatorname{BO}$  & \SI{3}{min} &  \SI{60}{min} & \SI{1}{min} &  \SI{20}{min} & \SI{10}{min} & \SI{200}{min}  \\
$\operatorname{cIDIR} + \operatorname{GS}$   & \SI{43}{min} & \SI{2}{sec}  & \SI{20}{min}  & \SI{2}{sec}   & \SI{90}{min} & \SI{2}{sec}  \\
\bottomrule
\end{tabular*}
\end{table}

\section{Conclusion}\label{Conclusion}

In this work, we introduced $\operatorname{cIDIR}$, a conditioned implicit neural representation (INR) for regularized deformable image registration. $\operatorname{cIDIR}$ is patient-specific and leverages an INR to model a continuous deformation vector field, enabling the integration of advanced regularization techniques that require higher-order gradients. By conditioning the activation functions of the INR on the regularization weighting factor, $\operatorname{cIDIR}$ allows for real-time hyperparameter optimization after training, eliminating the need for expensive retraining. Our experiments highlight $\operatorname{cIDIR}$'s accuracy, computational efficiency, and robustness across different regularization techniques and patient data.  Despite these advantages, $\operatorname{cIDIR}$ has limitations. Its patient-specific nature requires a dedicated training phase for each new subject, leading to long training times that may hinder its practical deployment, particularly in time-sensitive clinical settings. Another limitation is the assumption of a well-defined prior distribution for regularization parameters, which may not always align with the optimal settings for every case. Future work will focus on reducing training time to enhance practicality, exploring strategies to improve generalization across patients and datasets, and extending $\operatorname{cIDIR}$ to broader applications, such as multi-modal image registration. Finally, expanding $\operatorname{cIDIR}$ to handle multiple hyperparameters could allow for the efficient integration of diverse regularization techniques within a single training process, further strengthening diffeomorphism enforcement.



%
%
\bibliographystyle{splncs04}
\bibliography{paper}
\end{document}